\documentclass[10pt,twocolumn,letterpaper]{article}

\usepackage{iccv}
\usepackage{enumitem}
\usepackage{times}
\usepackage{epsfig}
\usepackage{graphicx}
\usepackage{amsmath}
\usepackage{amsmath, bm}
\usepackage{amssymb}
\usepackage{subfigure}
\usepackage{booktabs}
\usepackage{multirow}
\usepackage{url}

\usepackage[pagebackref=true,breaklinks=true,letterpaper=true,colorlinks,bookmarks=false]{hyperref}

\iccvfinalcopy 

\def\iccvPaperID{2251} 

\ificcvfinal\fi

\begin{document}

\title{Under-Display Camera Image Restoration with Scattering Effect}

\author{
	Binbin Song\textsuperscript{\rm 1}, 
	Xiangyu Chen\textsuperscript{\rm 1,2},
	Shuning Xu\textsuperscript{\rm 1}, and 
	Jiantao Zhou\textsuperscript{\rm 1\footnotemark[2]} \\
	\textsuperscript{\rm 1}State Key Laboratory of Internet of Things for Smart City\\
	Department of Computer and Information Science, University of Macau\\
	\textsuperscript{\rm 2}Shenzhen Institutes of Advanced Technology, Chinese Academy of Sciences\\
	{\tt\small \{yb97426, jtzhou\}@umac.mo, \{chxy95, rebeccaxu0418\}@gmail.com}
}


\maketitle
\renewcommand{\thefootnote}
{\fnsymbol{footnote}}
\footnotetext[2]{Corresponding author.}
\ificcvfinal\thispagestyle{empty}\fi

\begin{abstract}
The under-display camera (UDC) provides consumers with a full-screen visual experience without any obstruction due to notches or punched holes. However, the semi-transparent nature of the display inevitably introduces the severe degradation into UDC images. In this work, we address the UDC image restoration problem with the specific consideration of the scattering effect caused by the display. We explicitly model the scattering effect by treating the display as a piece of homogeneous scattering medium. With the physical model of the scattering effect, we improve the image formation pipeline for the image synthesis to construct a realistic UDC dataset with ground truths. To suppress the scattering effect for the eventual UDC image recovery, a two-branch restoration network is designed. More specifically, the scattering branch leverages global modeling capabilities of the channel-wise self-attention to estimate parameters of the scattering effect from degraded images. While the image branch exploits the local representation advantage of CNN to recover clear scenes, implicitly guided by the scattering branch. Extensive experiments are conducted on both real-world and synthesized data, demonstrating the superiority of the proposed method over the state-of-the-art UDC restoration techniques. The source code and dataset are available at \url{https://github.com/NamecantbeNULL/SRUDC}.
\end{abstract}

\section{Introduction}
\begin{figure}[t!]
	\centering
	\subfigure[UDC image]{
		\includegraphics[width=0.325\linewidth]{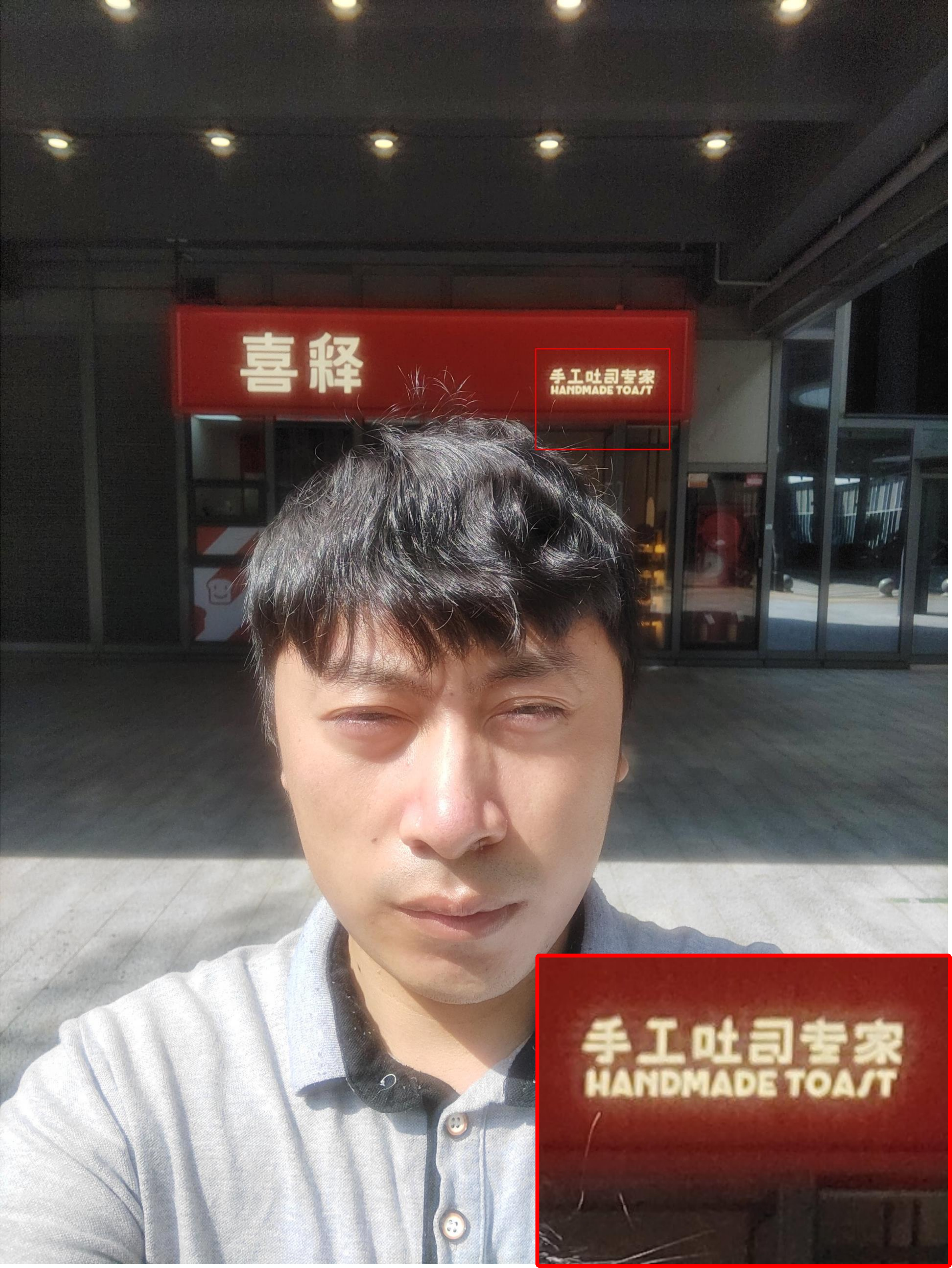}}
	\hspace{-2mm}
	\subfigure[DISCNet \cite{feng2021removing}]{
		\includegraphics[width=0.325\linewidth]{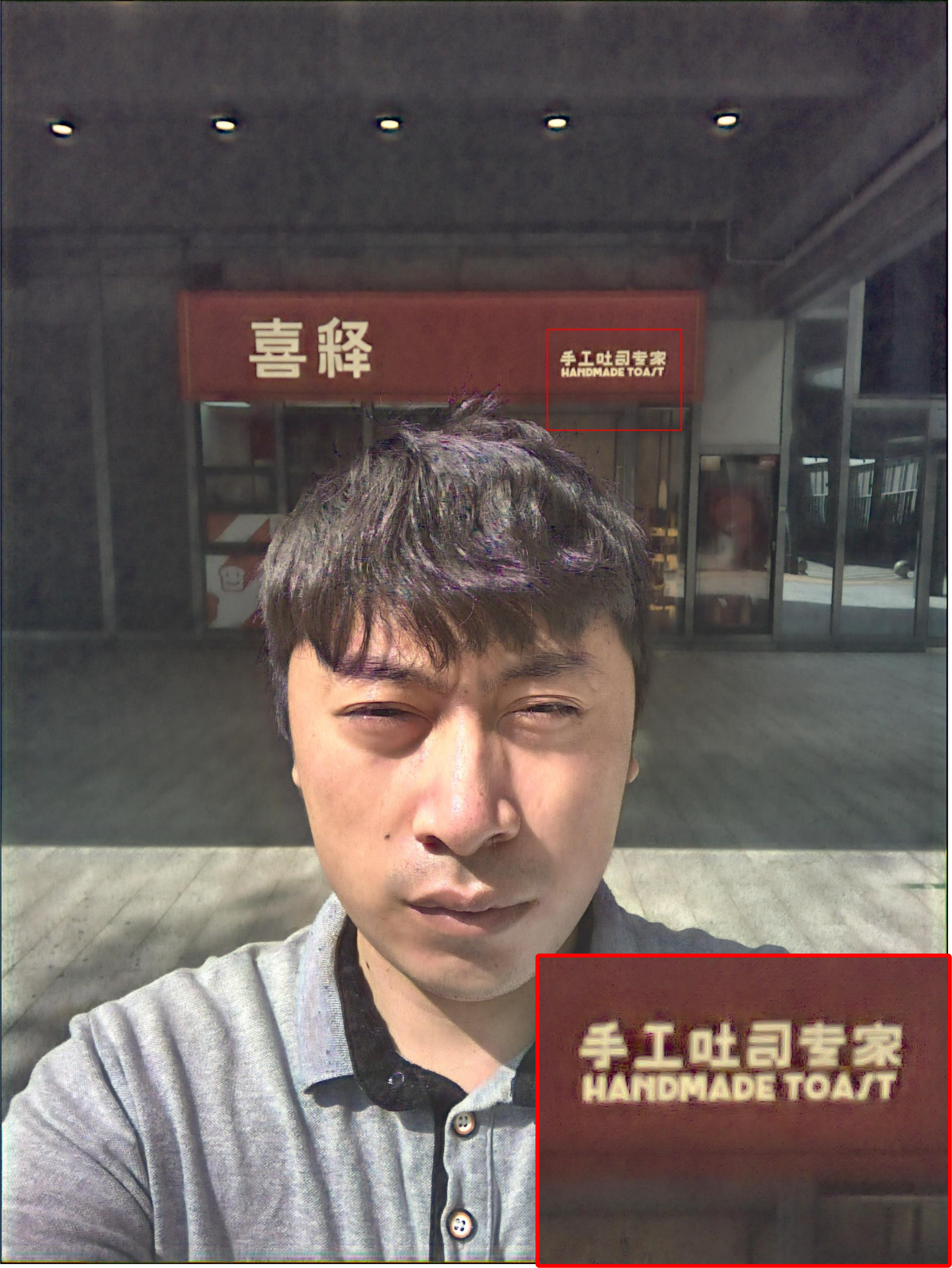}}
	\hspace{-2mm}
	\subfigure[DAGF \cite{sundar2020deep}]{
		\includegraphics[width=0.325\linewidth]{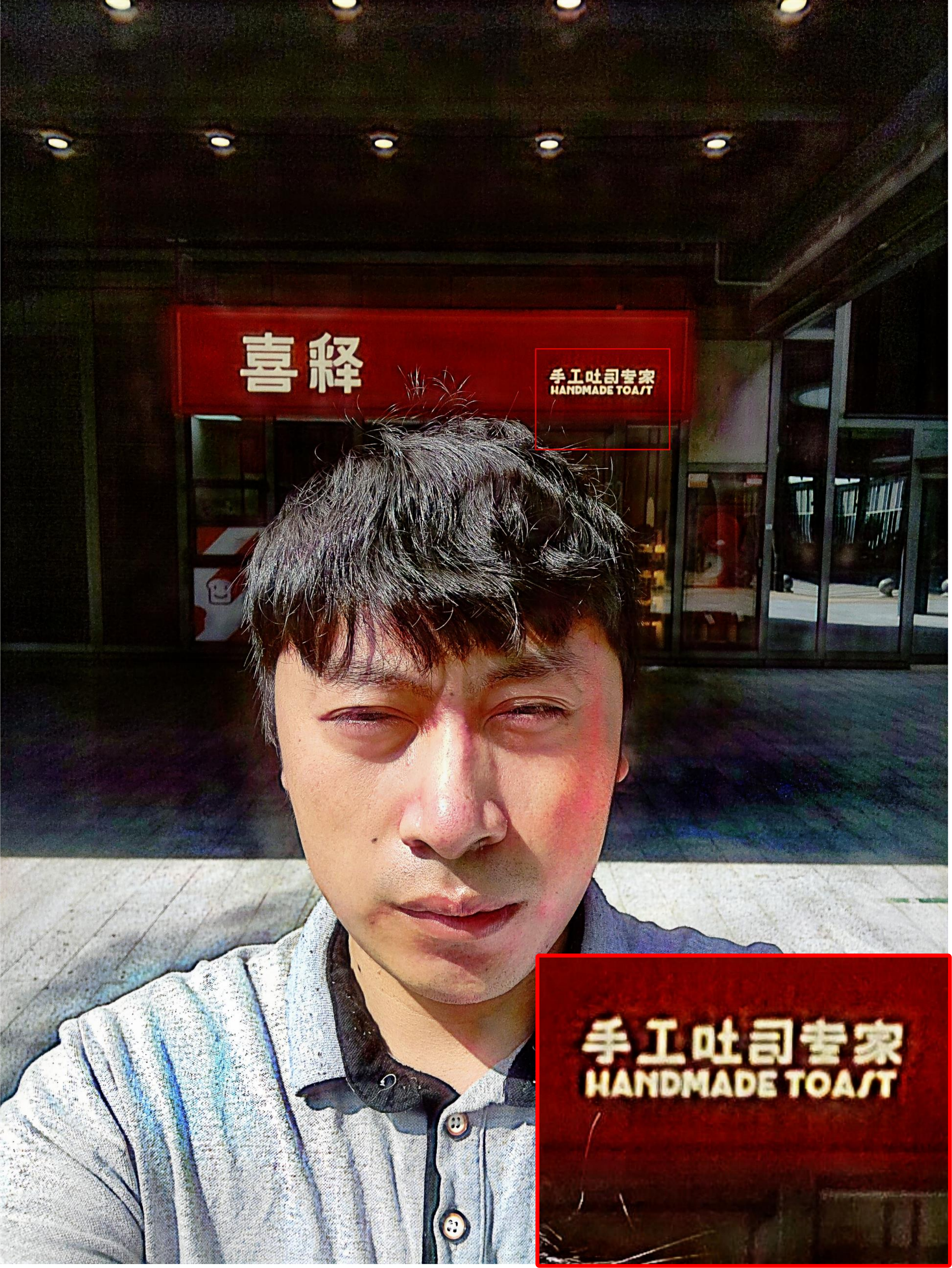}}
	\hspace{-2mm}
	\subfigure[UDCUnet \cite{Liu2022ECCV}]{
		\includegraphics[width=0.325\linewidth]{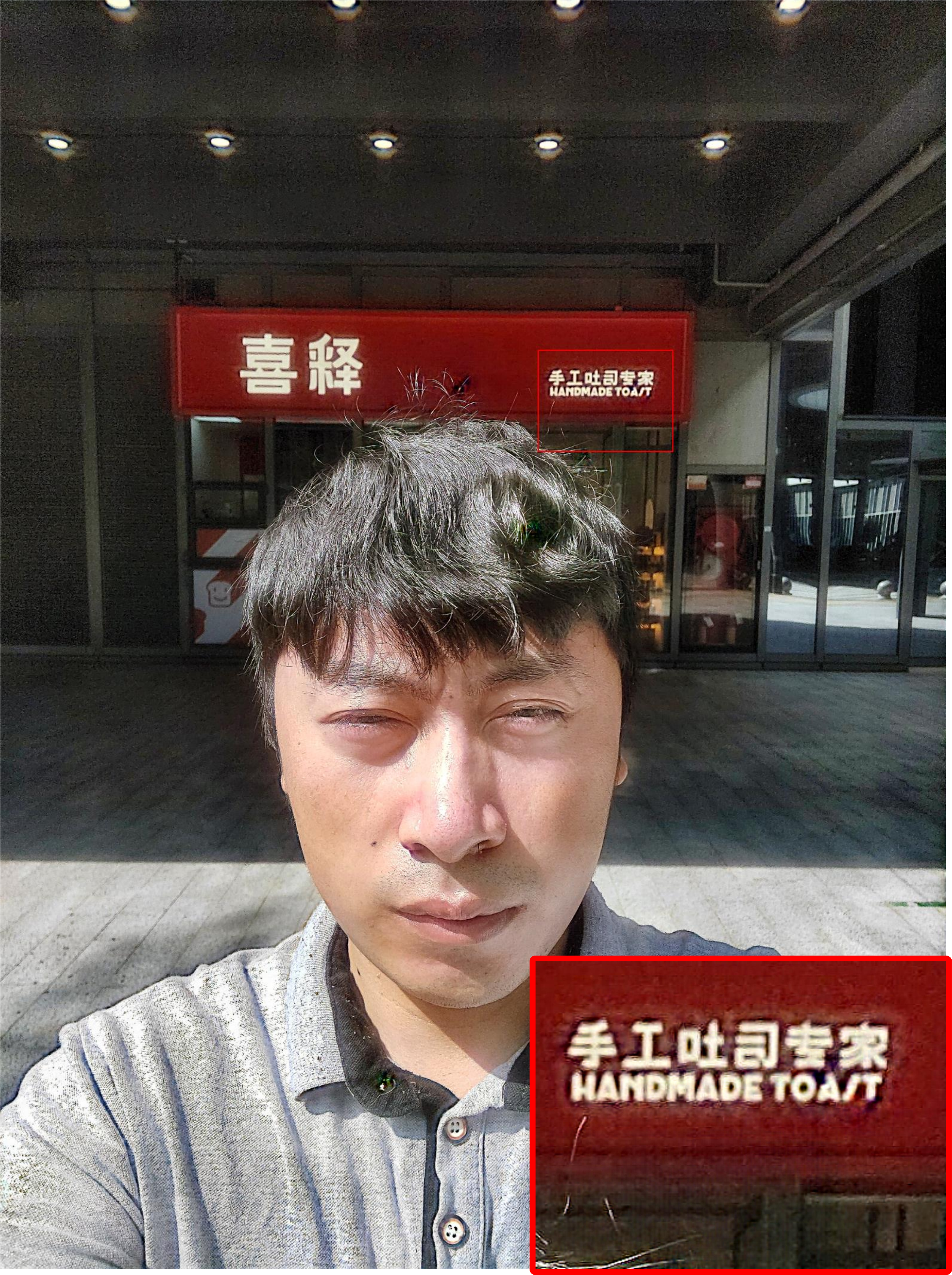}}
	\hspace{-2mm}
	\subfigure[BNUDC \cite{koh2022bnudc}]{
		\includegraphics[width=0.325\linewidth]{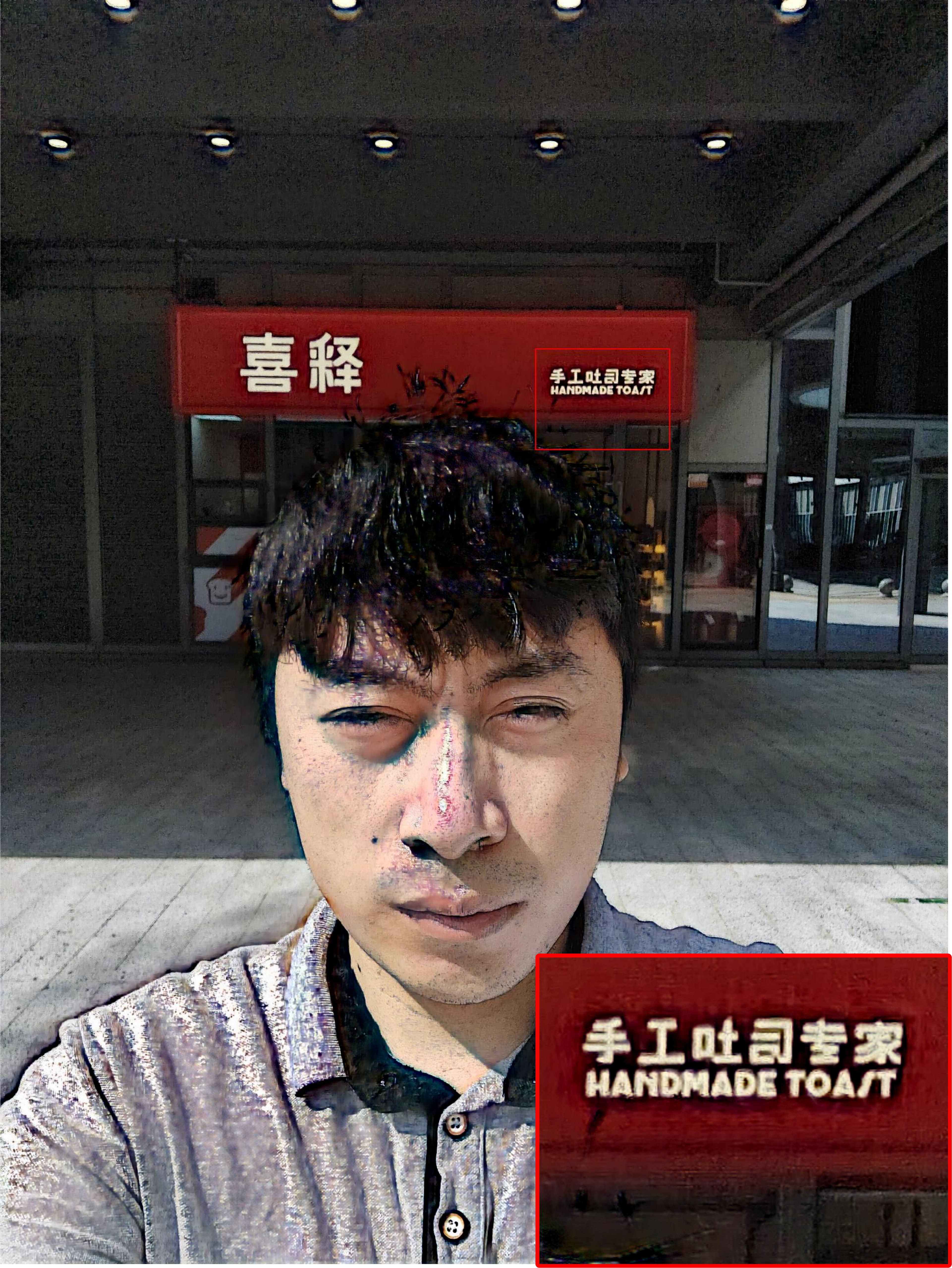}}
	\hspace{-2mm}
	\subfigure[Ours]{
		\includegraphics[width=0.325\linewidth]{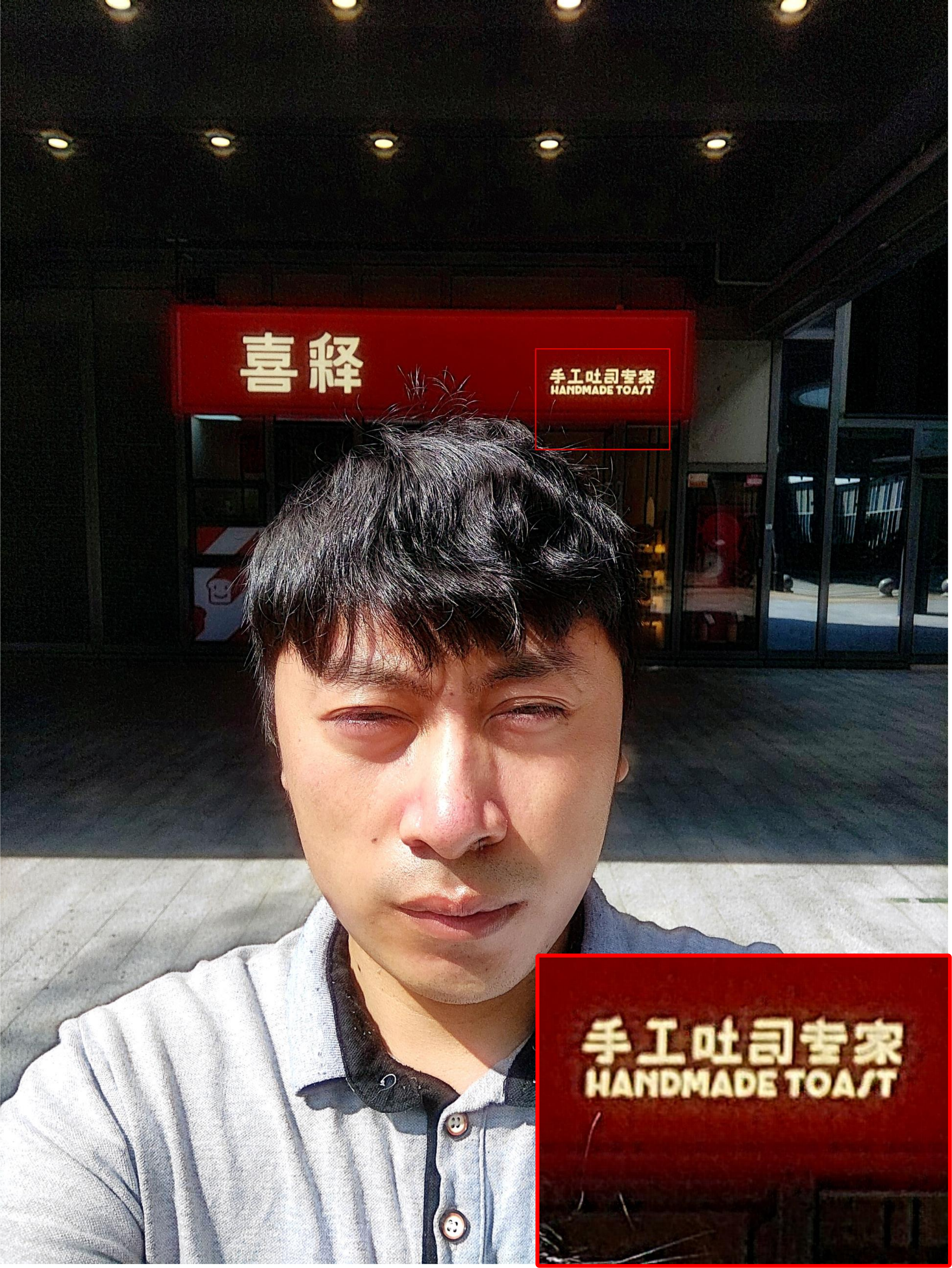}}
	\hspace{-2mm}
	\caption{An example of the UDC image restoration on a real-scene image. (a) Degraded UDC image, (b-e) Outputs of state-of-the-art methods w/o considering the scattering effect, (f) Our result, capable of removing  the haziness and contrast distortion, while restoring fine details. (Zoom in for a better view.)}
	\vspace{-3mm}
	\label{fig:intro}
\end{figure}
The adoption of under-display cameras (UDCs) has gained the popularity as a solution to avoid the inconvenience of notches and punched holes on the screen. This design meets the growing demand for full-screen hand-held devices and improves consumers' visual experience. However, the semi-transparent nature of the display covered on the camera inevitably introduces various types of the degradation, e.g., contrast distortion, blurring, haziness, noise, and diffraction artifacts around light sources, into the captured UDC images. To mitigate these issues, rapidly evolving UDC image restoration methods \cite{zhou2021image, feng2021removing, Kwon2021, koh2022bnudc, Gao2021optex, luo2022under, zhou2022modular, Melvin2020Transform, Conde2022realtime, sundar2020deep, feng2022mipi, Feng_2023_CVPR} in recent years are dedicated to recovering degraded images. Most of these methods focus on the degradation of high-frequency components, e.g., blurring, noise, and diffraction artifacts, while simply modeling the deterioration of low-frequency components as the pixel intensity attenuation. Whereas, the scattering effect of the display, which can result in the degradation such as contrast distortion and haziness, has been neglected in previous UDC restoration methods. Such a negligence may significantly limit the restoration performance and generalization of existing methods on real-scene images.

Essentially, the scattering effect is a common phenomenon in UDC imaging systems \cite{salehi2019recent}. When incident light enters the display, it travels through multiple stacked layers, e.g., the cathode, the substrate, etc., before reaching the camera sensor. During this process, collisions with small particles within these layers alter the propagation direction of some photons. Consequently, the total energy of the incident radiance is suppressed and part of the light is converted into scattered light, which generates contrast distortion and haziness. These types of degradation due to the scattering effect not only affect the visual quality of the captured UDC images, but also may cause the high-level vision systems to fail when UDC images are fed as the input. Surprisingly, as far as we know, the scattering effect has never been explicitly addressed in the UDC image restoration task. Due to the scattering effect, on one hand, there is a gap between the real-world and synthetic UDC images generated by the existing data synthesis pipeline. On the other hand, the restoration methods necessitate special designs for the scattering-induced degradation in UDC images.  These two reasons motivate us to take into account the scattering effect for better restoring the captured UDC images.   

In this work, we model the UDC scattering effect by decomposing the irradiance received by the camera into transmitted and scattered components and calculating them separately. Based on the proposed UDC scattering model, we enhance the existing image formation pipeline (IFP) to generate realistic UDC images with paired ground truths. We show that compared to the existing IFP, our enhanced method produces synthetic UDC images that are closer to real-captured ones. To restore UDC images with the scattering effect, we propose a two-branch deep network consisting of a scattering branch and an image branch. The scattering branch estimates the parameters of the scattering effect from degraded images, while the image branch recovers clear scenes guided by the scattering branch. In the scattering branch, we propose transposed self-attention blocks (TSABs) to leverage the global modeling capability \cite{zamir2022restormer, tu2022maxim} of the channel-wise self-attention (CSA). In the image branch, we use Convolutional Neural Network (CNN) as the backbone to exploit its local representation advantage \cite{Song2022MCDRNet, chen2021new}. To effectively combine the information from both branches, we devise a feature fusion module, which adaptively modulates the deep features with the global information from the scattering branch. Fig. \ref{fig:intro} presents an example of UDC image restoration on a real-scene image.     

In summary, our contributions are as follows:

\begin{itemize}[itemsep=-1pt, topsep=1.5pt]
	\item We propose a new UDC imaging model by explicitly considering the scattering effect, which causes unexpected haziness and contrast distortion. The image formation pipeline is then enhanced, generating more realistic UDC synthetic images.  
	\item Guided by the proposed UDC imaging model with the scattering effect, we specially design a dual-branch deep framework, which leverages global modeling capabilities of CSA and the local representation advantage of CNN, to restore the UDC image. 
	
	\item Extensive experiments demonstrate significant performance gains on both real-world and synthetic data. Compared with state-of-the-art methods, the visual quality of resulting images is dramatically improved.
\end{itemize}

\section{Related Works}\label{sec:related_works}
In this section, we first briefly introduce the works related to modeling the degradation process of UDC images. Then the recently developed UDC image restoration methods are reviewed.

\subsection{UDC imaging modeling}
Many previous works \cite{zhou2021image, feng2021removing, koh2022bnudc, 7403841, Kwon2021, park2013paper} have investigated the degradation of the UDC imaging. Zhou \textit{et al.}\cite{zhou2021image} first formulate the UDC imaging process with the classical convolution model as:
\begin{equation}\label{equ:ConvUDC}
	\mathbf{I}  = \gamma (\mathbf{B}  \ast \mathbf{k}) + \mathbf{n},
\end{equation}where $\mathbf{B}$ denotes the clean scene image, $\mathbf{I}$ denotes the degraded image. Here, $\mathbf{k}$ is the point spread function (PSF), $\mathbf{n}$ is the zero-mean noise, $\gamma$ is the intensity scaling factor, and $\ast$ represents the convolution operator. However, directly using this model in the limited dynamic range suppresses the diffraction artifacts around the saturated areas. To get closer to the real-world data, Feng \textit{et al.}\cite{feng2021removing} propose to migrate the convolution into the RAW domain with high dynamic range and measure the actual PSF of UDC in a smartphone for the data synthesis. Kwon \textit{et al.}\cite{Kwon2021} further consider the spatial variability of the PSF due to the incident angle of the light. To simulate complicated forms of the degradation with low spatial frequency, Koh \textit{et al.} \cite{koh2022bnudc} exploit a 3D transformation to substitute the single scaling factor $\gamma$.

Although the aforementioned works have improved the UDC degradation model, the convolution model with the single $\gamma$ cannot generate the widely-observed contrast distortion and haziness caused by the UDC scattering effect. This phenomenon motivates us to specifically model the light scattering process in the UDC imaging.

\subsection{UDC restoration methods}
The pioneering UDC image restoration method based on U-Net is proposed in \cite{zhou2021image}, as a blind deconvolution solution. To deal with the different PSFs in saturated regions, Feng \textit{et al.}\cite{feng2021removing} design a DynamIc Skip Connection Network (DISCNet) to estimate the latent clean image. Koh \textit{et al.} \cite{koh2022bnudc} propose a branched network for UDC image restoration (BNUDC), which simultaneously removes the high-frequency noise and the low-frequency degradation. To save memory for the restoration of high-resolution images, a Deep Atrous Guided Filter (DAGF) network is utilized by Sundar \textit{et al.} \cite{sundar2020deep}. Luo \textit{et al.} \cite{luo2022under} discover distinct statistical features of UDC and ground-truth images in HSV space. To fit the estimated images adaptively in H and S channels, they propose a cascaded curve estimation network for the enhancement of UDC images. 

While there have been some successful attempts on the UDC image restoration, existing methods have largely neglected the treatment of the scattering effect, which could severely affect the restoration performance on real-world UDC images.


\section{Enhancing IFP with UDC Scattering Model}\label{sec:degradation_model}
In this section, we first model the scattering phenomenon in UDC imaging systems by decomposing the irradiance received by the camera into transmitted and scattered components (Section \ref{sec:scattering_model}). Subsequently, we propose an enhanced approach for the UDC image formation pipeline (IFP) utilizing the scattering model for synthesizing UDC images (Section \ref{sec:imaging_model}). We demonstrate that our proposed method yields UDC images, which exhibit higher similarity to real-captured UDC images, compared to the existing UDC IFP.

\subsection{Modeling UDC Scattering Effect}\label{sec:scattering_model}
\begin{figure}[!t]
	\centering
	\includegraphics[width=0.8\linewidth]{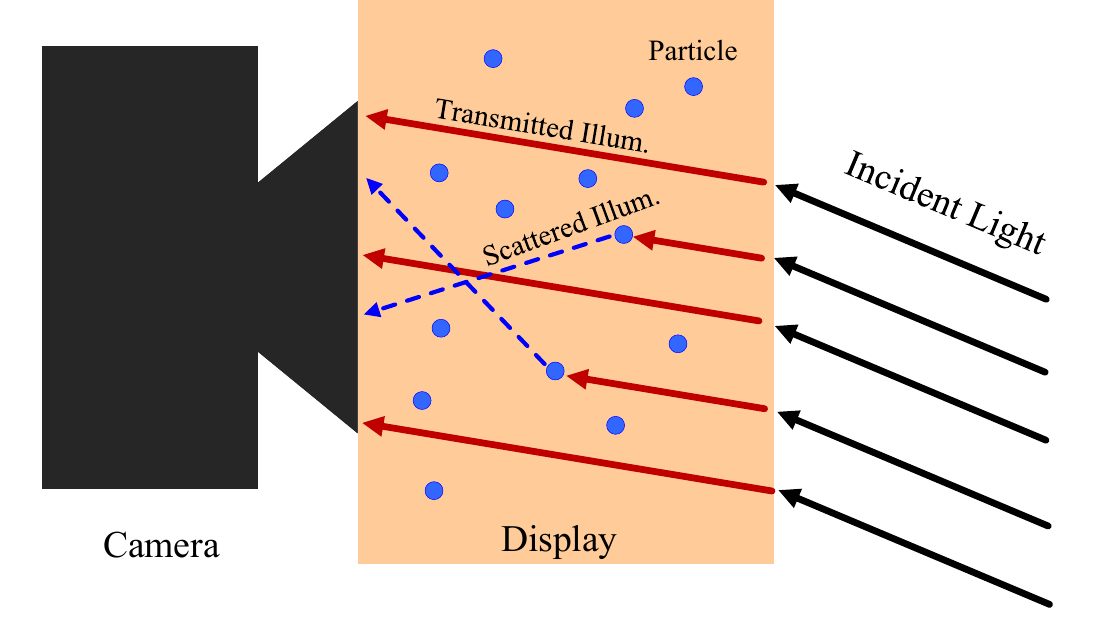}
	
	\caption{The schematic diagram of the UDC scattering model. The irradiance received by the camera can be decomposed into the transmitted illumination and scattered illumination.}
	\vspace{-3mm}
	\label{fig:scattering_model}
\end{figure}
In this study, we assume the incident radiance striking the display comes from a single light source, so as to facilitate the modeling of the UDC scattering effect in a simplified context. By exploiting the radiance accumulation, our modeling approach can be easily extended to the circumstance with multiple sources. In reality, the incident light can have a variety of sources, including sunlight, ground-reflected light, scene-reflected light, etc. Also, when the light source is considerably distant from the UDC, the incident light rays can be assumed as parallel light beams \cite{schuett2016perception}.

When the incident parallel light travels through the display, a fraction of photons collides with tiny particles inside the display, leading to the scattering phenomenon \cite{salehi2019recent}. The unaffected photons continue to pass through the display and eventually reach the camera sensor with attenuated amplitudes. A schematic illustration of the UDC scattering phenomenon is presented in Figure \ref{fig:scattering_model}. In our analysis, we treat the display in the UDC imaging system as a piece of homogeneous scattering medium with a constant scattering coefficient $\beta$. The UDC scattering effect can then be expressed as:
\begin{equation}\label{equ:scattering}
	\mathbf{L}(\mathbf{x}) = \mathbf{L}_t(\mathbf{x}) + \mathbf{L}_s(\mathbf{x}),
\end{equation}where $\mathbf{L}$ denotes the camera-captured light intensity, $\mathbf{L}_t$ is the transmitted illumination component and $\mathbf{L}_s$ is the scattered illumination component. Here, $\mathbf{x}$ represents the pixel location. Inspired by the scattering model used in image dehaze \cite{narasimhan2002vision, li2018benchmarking} and under-water image enhancement \cite{jaffe1990computer, li2021underwater}, we can write $\mathbf{L}_t$ and $\mathbf{L}_s$ in an explicit form:    
\begin{equation}\label{equ:scattering_comp}
	\begin{split}
		\mathbf{L}_t(\mathbf{x}) = \mathbf{L}_b(\mathbf{x}) \mathit{e}^{-\beta d(\mathbf{x})},\\
		\mathbf{L}_s(\mathbf{x}) = m(1-\mathit{e}^{-\beta d(\mathbf{x})}),
	\end{split}	
\end{equation}where $d(\mathbf{x})$ represents the routing path distance of the light ray within the display, and $m$ is the proportionality constant related to the incident radiance $\mathbf{L}_b$ (also known as the light intensity of the clean background image). For the light rays with the same incident angle, $d(\mathbf{x})$ approximates a fixed value for all the pixel positions. 

As will be discussed below, this UDC scattering model will be used to enhance the UDC IFP.

\subsection{Enhanced UDC IFP}\label{sec:imaging_model}
\begin{figure}[!t]
	\centering
	\includegraphics[width=1.0\linewidth]{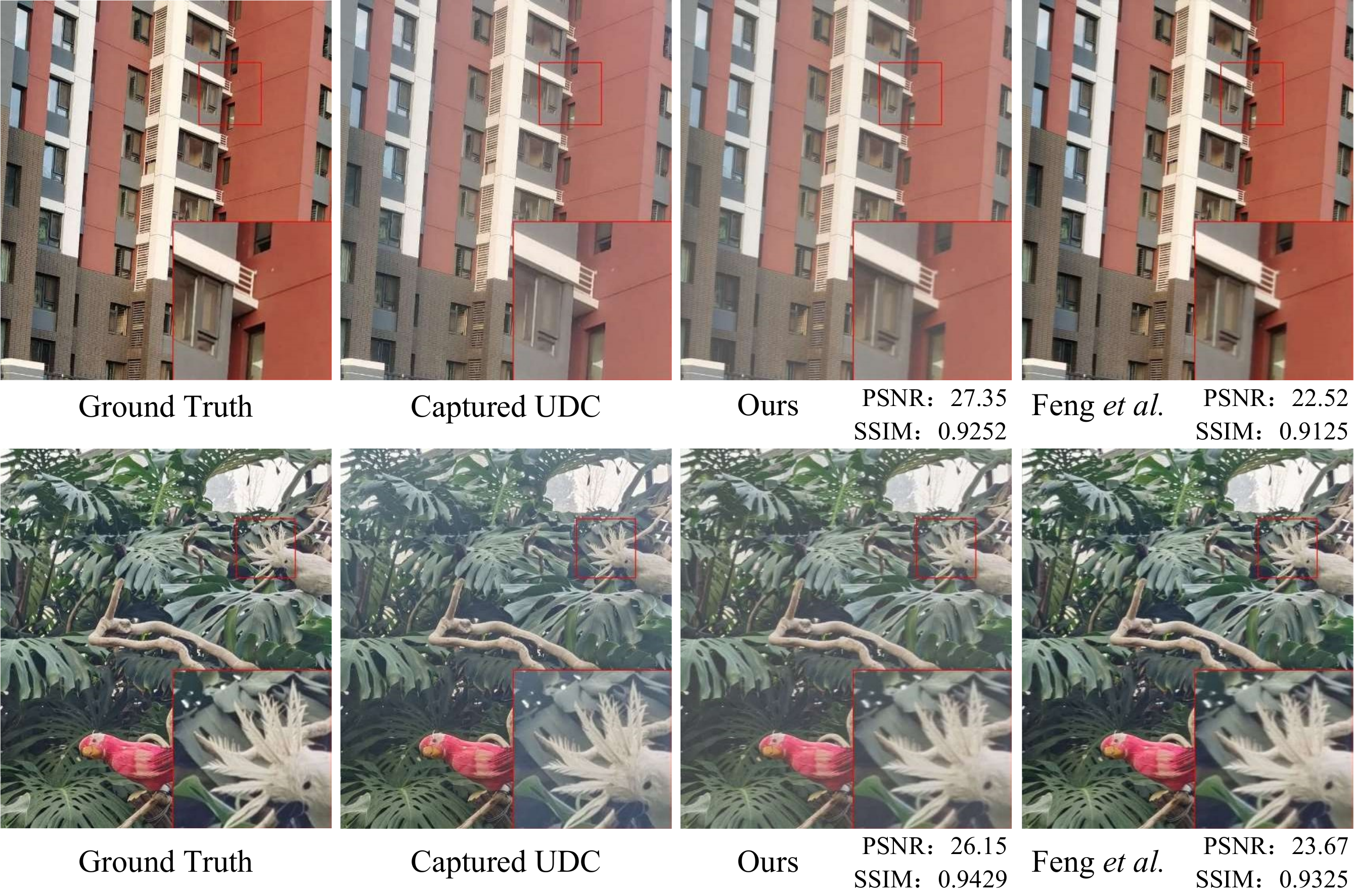}
	
	\caption{The comparison of synthetic images generated by our enhanced IFP and Feng \textit{et al.} \cite{feng2021removing}. Our enhanced IFP generates haziness and contrast distortion, which can also be observed in real-world captured UDC images. PSNR and SSIM are calculated by using the captured UDC images as the reference.}

	\vspace{-3mm}
	\label{fig:syn_images}
\end{figure}

\begin{figure*}[t!]
	\centering
	\includegraphics[width=0.9\textwidth]{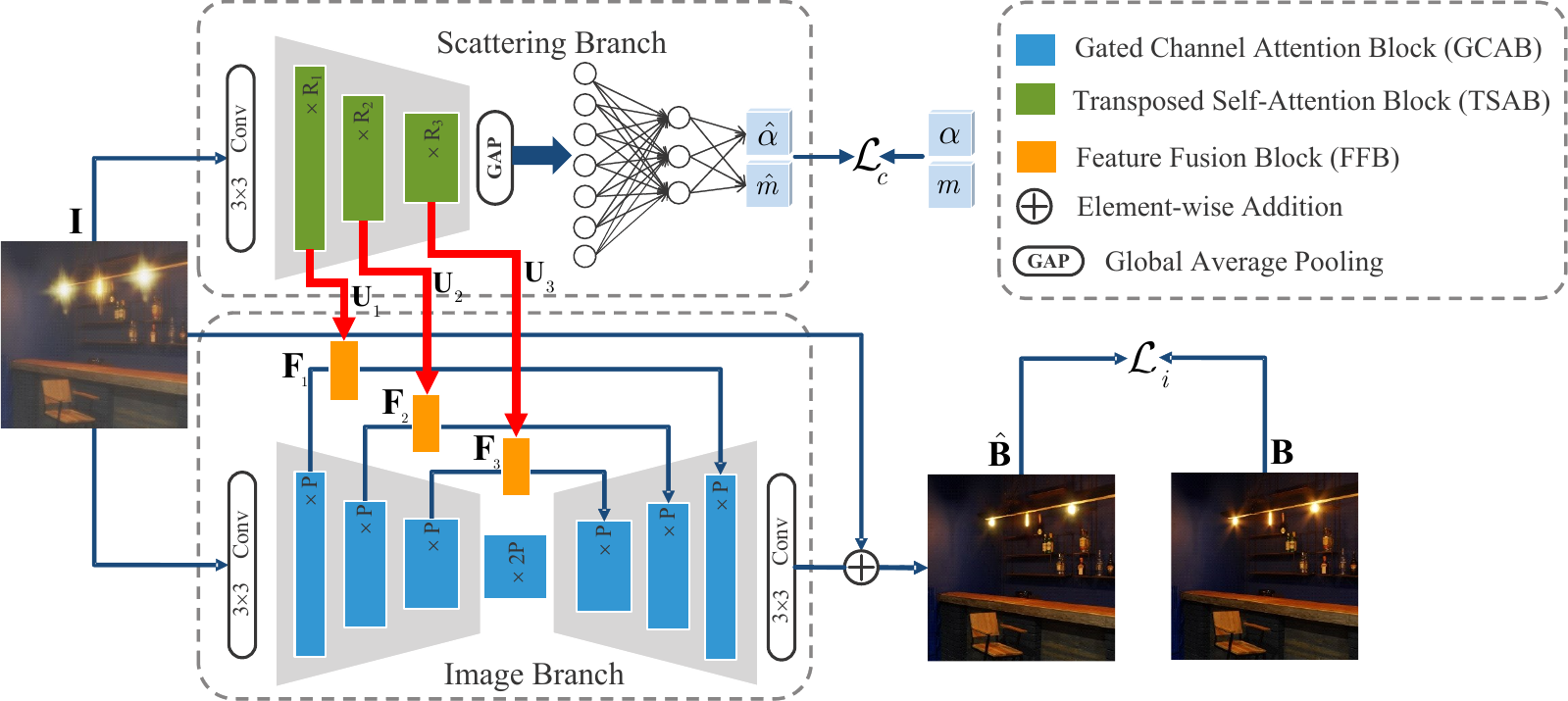}
	\caption{The overview of our proposed SRUDC for the UDC image restoration. The whole framework consists of two branches: the scattering branch and the image branch. The scattering branch estimates the parameters of the scattering effect from degraded images, while the image branch recovers clear scenes guided by the scattering branch.}
	\vspace{-3mm}
	\label{fig:framework}
	
\end{figure*}
Based on the proposed UDC scattering model, we now enhance the existing IFP to generate more realistic UDC images with paired ground truths for training the subsequent restoration networks. Specifically, we integrate the diffraction blur related to the PSF $\mathbf{k}$ and zero-mean noise $\mathbf{n}$ into \eqref{equ:scattering}. Our enhanced UDC IFP can then be defined as:  
\begin{equation}\label{equ:enhanced_IFP}
	\begin{split}
		\mathbf{I}(\mathbf{x}) = &TC\Big[(\mathbf{L}_t(\mathbf{x}) + \mathbf{L}_s(\mathbf{x})) \ast \mathbf{k} + \mathbf{n}\Big],
	\end{split}
\end{equation}where $\mathbf{x}$ is the pixel position, $\mathbf{I}$ is the degraded UDC image, and $TC(\cdot)$ denotes the tone mapping and clipping functions. Here, the purpose of introducing $TC(\cdot)$ is to convert the image intensities into the low-dynamic-range space after generating the anisotropic diffraction blur in the high-dynamic-range space. Substituting \eqref{equ:scattering_comp} into \eqref{equ:enhanced_IFP} yields:
\begin{equation}\label{equ:detailed_IFP}
	\begin{split}
		\mathbf{I}(\mathbf{x}) = TC\Big\{[\mathbf{L}_b(\mathbf{x}) \mathit{e}^{-\beta d(\mathbf{x})} + m(1-\mathit{e}^{-\beta d(\mathbf{x})})] \ast \mathbf{k} + \mathbf{n}\Big\}.
	\end{split}
\end{equation}where $\mathbf{L}_b(\mathbf{x})$ is the light intensity of the background $\mathbf{B}$. As $d(\mathbf{x})$ approximates a constant for all $\mathbf{x}$s and both $\mathbf{k}$ and $\mathbf{n}$ are assumed to be independent with $\mathbf{x}$, the above UDC scattering model can be seen as a global transformation. By introducing $\alpha = \mathit{e}^{-\beta d}$ and omitting $\mathbf{x}$, the enhanced IFP can be further simplified as:   
\begin{equation}\label{eq:IFP}
	\begin{split}
		\mathbf{I} = TC\Big\{[\alpha \mathbf{L}_b + m(1-\alpha)] \ast \mathbf{k} + \mathbf{n}\Big\}.
	\end{split}
\end{equation} Regarding the parameter $m$, which is the proportionality constant related to the incident radiance $\mathbf{L}_b$, we set it as the mean of the gray-scale $\mathbf{B}$ to maintain the average brightness. Such a setting is reasonable since the under-display photography generally could not influence the distribution of the value (V) channel in HSV space \cite{luo2022under}. Also, $\alpha$ is a constant empirically sampled in the interval $[0.6, 0.9]$.   

To justify the rationality of our enhanced IFP, we compare it with the existing data synthesizing method \cite{feng2021removing} on a small real dataset \cite{luo2022under}. As demonstrated in Fig. \ref{fig:syn_images}, our generated UDC images exhibit a closer visual resemblance to the real-captured images; also with significantly higher PSNR and SSIM values. To further illustrate the superiority of our enhanced IFP, we conduct a statistical analysis of the entire synthetic dataset in the HSV color space. We find that the distributions of the hue (H) and saturation (S) channels in our synthetic dataset are similar to those of the real-captured dataset.  The histogram representations of statistical properties can be found in the supplementary material.

\section{Scattering Removal Network for UDC Image Restoration}\label{sec:network}
To specifically address the problem of the scattering effect in the UDC image restoration, we propose a Scattering Removal network for the UDC image restoration (SRUDC). Based on our enhanced UDC IFP, we notice that accurate estimates of $\alpha$ and $m$ are rather beneficial to the UDC image recovery. To this end, we purposely design a scattering branch in SRUDC that estimates these parameters to guide the restoration process of the background image in the image branch. The overview of the framework is shown in Fig. \ref{fig:framework}. Given the degraded UDC image $\mathbf{I}$, our target is to restore the clean background image $\mathbf{B}$. Our SRUDC framework consists of two branches, namely the scattering branch and the image branch. Concretely, the scattering branch estimates $\alpha$ and $m$ from $\mathbf{I}$ (Section \ref{sec:scattering_branch}), while the image branch restores $\mathbf{B}$ with the guidance of the scattering branch (Section \ref{sec:image_branch}). To further enhance the feature representation of the image branch, we design a new feature fusion block (FFB) that effectively combines the global information from the scattering branch (Section \ref{sec:FFB}).

\subsection{Scattering Branch}\label{sec:scattering_branch}
The scattering branch is designed to estimate $\alpha$ and $m$, which numerically represent the scattering effect in the UDC imaging system. Recall from (\ref{eq:IFP}) that these two coefficients determine the global transformation of the clean image due to the UDC scattering effect. As a result, the accurate estimation of $\alpha$ and $m$ requires the network to maintain the capability of modeling spatially invariant global information. To this end, inspired by the recent success of channel-wise self-attention \cite{zamir2022restormer, yan2021channel, Wang22CDGAN}, we develop the Transposed Self-Attention Block (TSAB), which captures long-range dependencies in the channel dimension.  Furthermore, we employ global average pooling (GAP) to extract the useful global statistics of the features, and then calculate the conditional vector of $\hat{\alpha}$ and $\hat{m}$ through a two-layer multi-layer perceptron (MLP).



The detailed structure of the scattering branch is demonstrated in the upper part of Fig. \ref{fig:framework}. Given a degraded UDC image $\mathbf{I}$, we apply a $3\times3$ convolution layer to extract the shallow features. The extracted features then pass through a three-level encoder built upon several TSABs to generate the set of deep features  $\mathbf{U}_o (o = 1, 2, 3)$, where $o$ is the level index. The numbers of TSABs in each level are denoted as $\{R_1, R_2, R_3\}$. We downscale features twice to gradually reduce the spatial size and increase the number of channels. Eventually, the estimated $\hat{\alpha}$ and $\hat{m}$ are obtained by:
\begin{equation}
 \vspace{-1mm}
	(\hat{\alpha}, \hat{m}) = W_{2} \odot g(W_{1} \odot GAP(\mathbf{U}_3)),
\end{equation}where $GAP(\cdot)$ means the global average pooling layer, $W_{1}$ and $W_{2}$ are parameters of the two-layer MLP, $g(\cdot)$ denotes a Sigmoid function, and $\odot$ represents the Hadamard product. Also, the features $\mathbf{U}_o$s are further transformed to guide the reconstruction of the clean scenes in the image branch.

\begin{figure}[t!]
	\centering
	\includegraphics[width=0.9\linewidth]{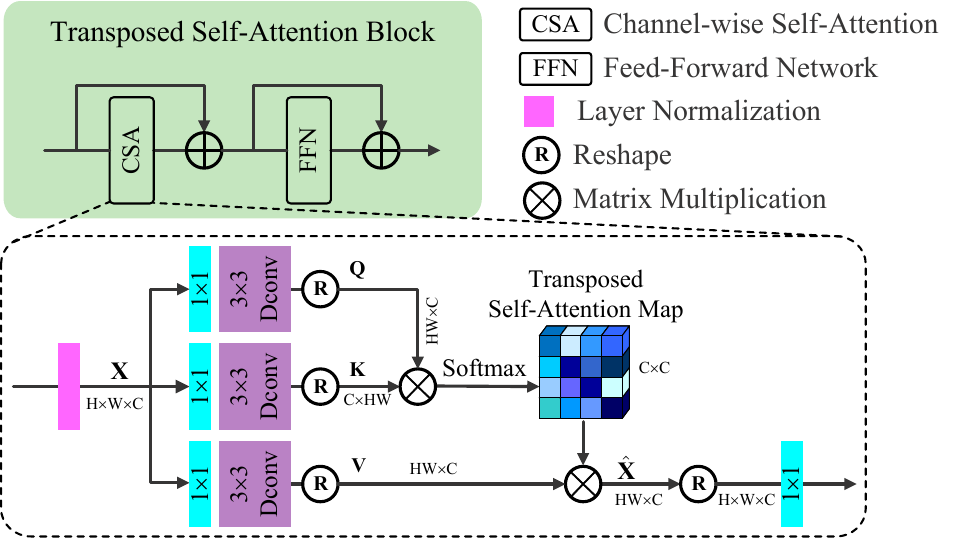}
	\caption{The structure of TSAB. }
	\vspace{-5mm}
	\label{fig:TSAB}
\end{figure}

Regarding the TSAB, which is an essential building block in the scattering branch, its architecture is shown in Fig. \ref{fig:TSAB}. Given the input features, to capture the global information through long-distance dependencies in the channel dimension and significantly reduce the storage consumption, we compute CSA for the modulation of the feature residue. From a layer normalized tensor $\mathbf{X}\in \mathbb{R}^{H \times W \times C}$, we use the combinations of $1\times1$ convolution and $3\times 3$ depth-wise convolution followed by the reshaping to generate $\{\mathbf{Q}, \mathbf{K}, \mathbf{V}\} \in \mathbb{R}^{HW \times C}$. The CSA modulation is then defined as:   
\begin{equation}
 \vspace{-1mm}
	\hat{\mathbf{X}} = softmax(\mathbf{Q} \mathbf{K}^{T}) \mathbf{V},
\end{equation}where $\hat{\mathbf{X}}$ means channel-wise modulated features. Then we reshape $\hat{\mathbf{X}}$ into a dimension $\mathbb{R}^{H \times W \times C}$ for further processing. After the modulation by the CSA, the feed-forward network is added in TSAB to introduce the nonlinear activation, further enhancing the capability of CSA.

\subsection{Image Branch}\label{sec:image_branch}
With the guidance of $\mathbf{U}_o$s from the scattering branch, the image branch is proposed to generate the estimated background $\hat{\mathbf{B}}$. We leverage CNN as the backbone to exploit its local representation capability. As shown in the bottom part of Fig. \ref{fig:framework}, $\mathbf{I}$ is first transformed into the feature space through a $3\times3$ convolution layer. Then, a symmetric U-shape architecture reconstructs the background scenes in the feature space. The encoder and the decoder consist of three levels, built based on our proposed Gated Channel Attention Block (GCAB). To assist the recovery process, the features from the encoder, denoted as $\mathbf{F}_e (e = 1, 2, 3)$, are modulated by $\mathbf{U}_o$ and then concatenated with the decoder features via skip connections \cite{ronneberger2015u}.

\begin{figure}[t!]
	\centering
	\includegraphics[width=0.8\linewidth]{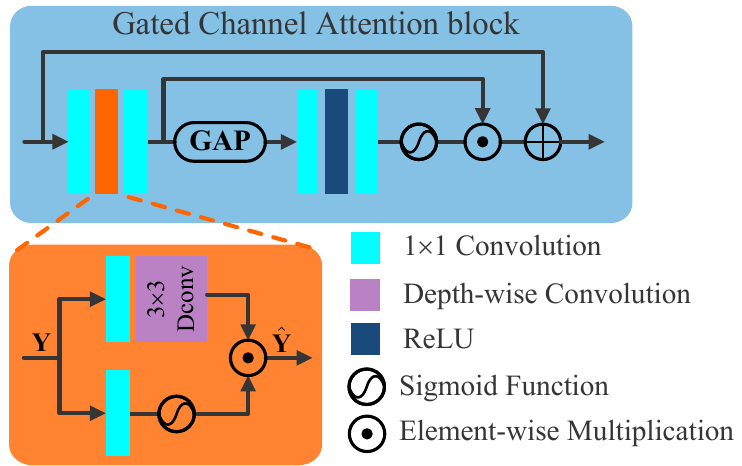}
	\caption{The structure of the proposed GCAB. }
	\vspace{-5mm}
	\label{fig:GCAB}
\end{figure}

To better reconstruct the spatially variant UDC image content, we design GCAB as the basic building block of the encoder and the decoder in the image branch. The structure of GCAB is depicted in Fig. \ref{fig:GCAB}. GCAB is based on the widely used residual channel attention block \cite{Zhang2018RCAB, zamir2021multi}. To enhance the spatial representation capability of the feature maps, we introduce the gating mechanism \cite{chen2022simple, song2022rethinking}, which adaptively activates the feature maps. As shown in the orange block in Fig. \ref{fig:GCAB}, given an input tensor $\mathbf{Y}$, the gating operator is defined as:
\begin{equation}
	Gating(\mathbf{Y}) = W_{dw} W_{pw}^{1} \mathbf{Y} \odot g(W_{pw}^{2}  \mathbf{Y}),
\end{equation}where $W_{pw}^{1}$ and $W_{pw}^{2}$ respectively represent the weight matrices of two $1\times1$ convolution layers, and $W_{dw}$ is the parameter of the $3\times3$ depth-wise convolution. By integrating the gate mechanism and the channel attention, we can adaptively modulate the feature in both the spatial and the channel dimensions for the restoration of UDC images.

\subsection{Feature Fusion Block}\label{sec:FFB}
To implicitly guide the estimation of $\hat{\mathbf{B}}$ in the decoder of the image branch, we recalibrate $\mathbf{F}_e$ with the scattering effect information embedded in  $\mathbf{U}_o$, via Feature Fusion Blocks (FFBs). Since $\mathbf{F}_e$ and $\mathbf{U}_o$ respectively encode the global and local information of the input, directly concatenating or element-wisely adding them may cause ambiguity. To solve this challenge, we devise the FFB to adaptively modulate $\mathbf{F}_e$. According to the formulation of the UDC scattering model (see \eqref{eq:IFP}), we propose to utilize the affine transformation in FFB. Inspired by spatial feature transformation (SFT) \cite{wang2018recovering, Chen2021HDRUnet}, we compute the coefficient and offset of the affine transformation from $\mathbf{U}_o$. The illustration of the design of FFB is put in the supplement. Given $\mathbf{F}_e$ and $\mathbf{U}_o$ with $e=o$, the fusion weights $\{\mathbf{v}, \mathbf{w}\}$ can be obtained through the combination of two $1\times1$ convolution layers and a leakey ReLU function. Then, the output feature $\mathbf{F}_o$ can be generated by:
\begin{equation}
	\mathbf{F}_o = \mathbf{v} \odot \mathbf{F}_e + \mathbf{w}.
\end{equation}

\subsection{Training Objectives}
We apply the supervision on the estimated $\{\hat{\alpha}, \hat{m}\}$ and the restored $\hat{\mathbf{B}}$. Our framework is trained in an end-to-end manner, with the following objective function: 
\begin{equation}
	\mathcal{L} = \omega_{c} \mathcal{L}_{c} + \omega_{i} \mathcal{L}_{i},
\end{equation}
where $\mathcal{L}_{c}$, and $\mathcal{L}_{i}$ respectively represent training losses for the scattering branch and the image branch. Here, the weighting parameters $\omega_{c}$ and $\omega_{i}$ that balance the relative importance of the scattering and image branches are empirically set as $\omega_{c}=0.1$ and $\omega_{i}=1$. For the scattering branch, we use mean square error (MSE) as the criterion, while for the image branch, the training loss is defined in a $l_1$ form. Specifically, we have:  
\begin{equation}
    \begin{split}
	\mathcal{L}_{c} = &\mathcal{L}_{MSE}(\hat{\alpha}, \alpha) + \mathcal{L}_{MSE}(\hat{m}, m), \\
    \mathcal{L}_{i} = &\mathcal{L}_{l_1}(\hat{\mathbf{B}}, \mathbf{B}).
    \end{split}
\end{equation}

\section{Experimental Results} \label{sec:experiments}
We now describe some details regarding the implementation of SRUDC. For the numbers of TSABs in the scattering branch, we set $\{R_1=2, R_2=3, R_3=4\}$, while for the number of GCABs in the image branch, we set $P=16$. The proposed SRUDC is implemented with Pytorch \cite{paszke2017automatic} framework and trained for 800 epochs on four NVIDIA V100 GPUs. During the training process, the images are randomly cropped into $256 \times 256$ patches, and the mini-batch size is set to 64. The AdamW optimizer \cite{loshchilov2017decoupled} is used to update learnable parameters. We use the cosine annealing strategy \cite{loshchilov2016sgdr} to adjust the learning rate, which is initialized to be $4\times 10^{-4}$ and gradually decreases to $4\times 10^{-6}$. 

\subsection{Training and Testing Datasets}
\vspace{-1mm}
The proposed SRUDC is trained with a synthetic dataset. The degraded image $\mathbf{I}$ is obtained by using \eqref{eq:IFP}. For the PSF $\mathbf{k}$ in \eqref{eq:IFP}, we collect 10 kernels in total, including 9 measured ones with ZTE Axon 20 phone \cite{feng2021removing} and a real-world one with Transparent-OLED \cite{Yang2021TPAMI}. With these PSF kernels, we could generate 20160 training image pairs from 2016 image patches, which are cropped with the size $800 \times 800$ from the HDR images in HDRI Haven dataset \cite{HDRI2023}.


Regarding testing datasets, we exploit the combination of synthetic and real-world data for the evaluation. Similar to the synthetic training dataset, we select 360 cropped image patches from HDRI Haven dataset to generate the synthetic testing set. The real testing set is formed by three parts, including \texttt{UDC-ZTE} \cite{feng2021removing} with 30 high-resolution UDC images taken by ZTE Axon 20 phone, \texttt{UDC-UAV} \cite{feng2021removing} with 162 images captured from the aerial video, and \texttt{UDCIE} \cite{luo2022under} with 12 images taken in the wild environment. Hence, in total, we have 3600 synthetic and 204 real-world testing images for evaluating our method and the existing competitors.     

\subsection{Quantitative results}
\vspace{-1mm}
\begin{table}[t]
	\setlength\tabcolsep{5.5pt}
	\centering
	\caption{Quantitative comparison on the synthetic dataset. The best results are in \textbf{bold} and the second best results are \underline{underlined}.}
	\label{tab:syn_comp}
	\scalebox{0.88}{
		\begin{tabular}{l|ccc|cc}
			\toprule[1pt]
			Method                        & PSNR  & SSIM  & LPIPS & \begin{tabular}[c]{@{}c@{}}Params\\  (M)\end{tabular} & \begin{tabular}[c]{@{}c@{}}MACs \\ (G)\end{tabular}  \\
			\midrule
			DISCNet \cite{feng2021removing}  & 25.54 & 0.9246 & 0.1508 & 3.442 & 272.33 \\
			DAGF \cite{sundar2020deep}         & 27.12 & 0.9230 & 0.1024 & \textbf{1.117} & \textbf{45.89} \\
			DWFormer \cite{zhou2022modular}  & 28.85 & 0.9392 & 0.1366 & 1.447 & 131.84 \\
			UDCUNet \cite{Liu2022ECCV}     & 31.33 & 0.9759 & 0.0459 & \underline{1.406} & 402.16 \\
			BNUDC \cite{koh2022bnudc}        & 34.91 & 0.9777 & \underline{0.0331} & 4.575 & 317.38 \\
			SRUDC-light                      	           & \underline{35.56} & \underline{0.9804} & 0.0365 & 3.105 & \underline{106.79} \\
			SRUDC-full                                    & \textbf{36.02} & \textbf{0.9833} & \textbf{0.0322} & 8.734 & 240.36 \\
			\bottomrule[1pt]
	\end{tabular}}
    
\vspace{-5mm}
\end{table}

\begin{figure*}[!h]
	\centering
	\includegraphics[width=1.0\linewidth]{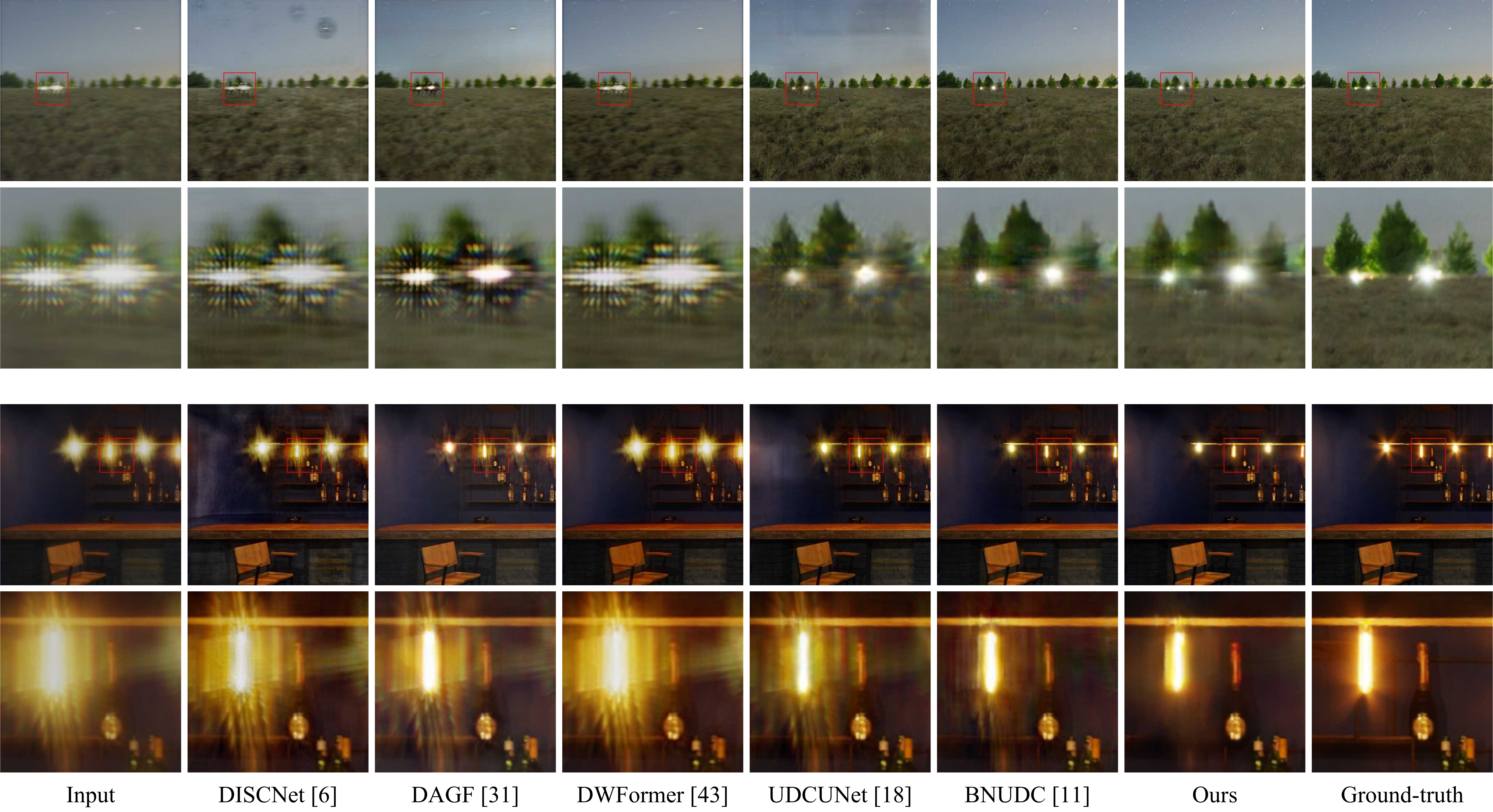}
	\caption{Visual comparison on synthetic images. Our method removes the scattering-induced haziness and contrast distortion spread over the entire image. Meanwhile, the diffraction artifacts around the light sources are greatly suppressed.}
	\vspace{-3mm}
	\label{fig:result_syn}
\end{figure*}

The quantitative results tested on the synthetic data are reported in Table \ref{tab:syn_comp}. We provide two versions of SRUDC: a full version and a lightweight version denoted as SRUDC-full and SRUDC-light, respectively. The numbers of the feature channel for the three levels of the encoder-decoder in SRUDC-full and SRUDC-light are respectively set as $\{24, 48, 96\}$ and $\{16, 32, 64\}$. We compare the restoration performance of SRUDC with that of five state-of-the-art UDC image restoration methods, including DISCNet \cite{feng2021removing}, DAGF \cite{sundar2020deep}, DWFormer \cite{zhou2022modular}, UDCUNet \cite{Liu2022ECCV}, and BNUDC \cite{koh2022bnudc}. For a fair comparison, we also retrain these models on our training data with the default parameters set as suggested, and choose the better results between pretrained and retrained models for the comparison. This is certainly advantageous to the competing methods. We adopt PSNR, SSIM \cite{zhou2004SSIM}, and LPIPS \cite{Zhang_2018_LPIPS} as evaluation metrics. The numbers of parameters and multiply-accumulation operations (MACs) are used to compare the model complexity.  As tabulated in Table \ref{tab:syn_comp}, SRUDC-full outperforms the second-best method BNUDC by a large margin, i.e., 1.11dB in PSNR, 0.0056 in SSIM, and 0.0009 in LPIPS. For the model complexity, SRUDC-light achieves the second-lowest computation overhead with 106.79G MACs and a comparable number of parameters with existing SOTA methods. Meanwhile, SRUDC-light still maintains a performance gain of 0.65dB in PSNR over BNUDC.




\subsection{Qualitative Results}
\begin{figure*}[!h]
	\centering
	\includegraphics[width=1.0\linewidth]{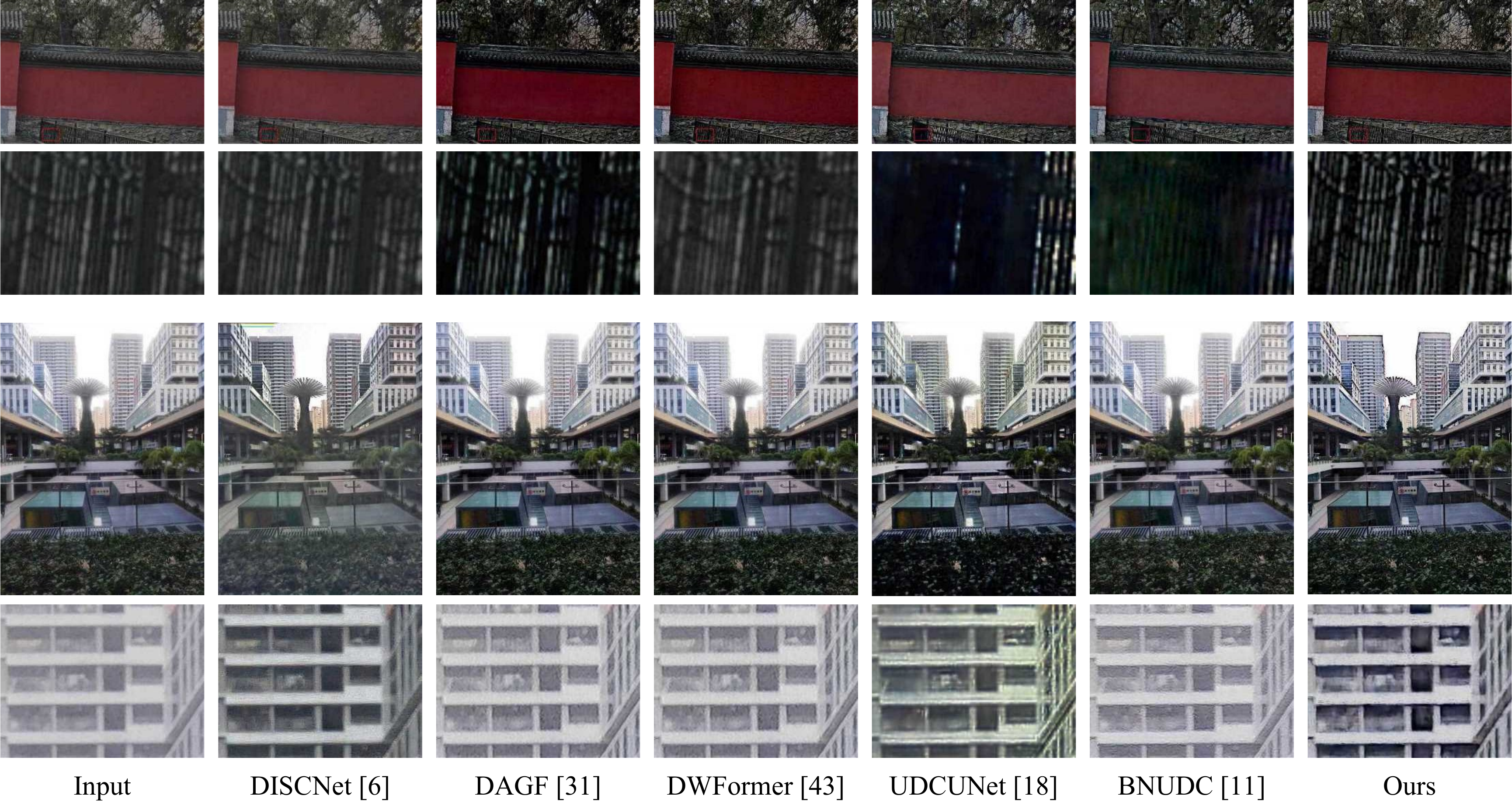}
	\caption{Visual comparison on real-world images. Our method achieves the best perceptual quality for the reconstruction of the textual regions, e.g., repeated fences and building structures.}
	
	\label{fig:result_real}
 \vspace{-3mm}
\end{figure*}

\vspace{-1mm}
We now compare the qualitative results of our method and the competing approaches on both synthetic and real-world UDC images. Fig. \ref{fig:result_syn} presents the results on two synthetic examples. It can be seen that our method is capable of removing the scattering-induced haziness and contrast distortion spread over the entire image, and greatly suppressing the diffraction artifacts around the light sources. We also give the visual comparison of different methods on real-world datasets in Fig. \ref{fig:result_real}. As can be observed, our SRUDC achieves the best perceptual quality, especially in terms of the reconstruction of the textual regions, e.g., repeated fences and building structures. In addition, SRUDC avoids color shifts as UDCUnet. More qualitative comparisons can be found in the supplementary file.


\subsection{Ablation study}
\vspace{-1mm}

\begin{table}[t]
	\centering
	\caption{Impact of the scattering branch. }
	\label{tab:scattering_branch}
	\begin{small}
		\begin{tabular}{l|ccc}
			\toprule[1pt]
			Model & PSNR & SSIM  & LPIPS \\
			\midrule
			w/o the supervisions of $m$ and $\alpha$& 35.13 & 0.9817 & 0.0347 \\
			w/o the entire scattering branch & 34.66 & 0.979 & 0.0394\\
			w/ CNN scattering branch & 35.02 &0.9798  & 0.0336 \\
			SRUDC-full & \textbf{36.02} & \textbf{0.9833} & \textbf{0.0322}\\
			\bottomrule[1pt]
		\end{tabular}
	\end{small}
\vspace{-5mm}
\end{table}

\noindent\textbf{\underline{Scattering Branch:}} To demonstrate the benefits of the scattering branch, we build the following ablation experiments. We first evaluate the importance of including the supervision of $m$ and $\alpha$. As shown in Table \ref{tab:scattering_branch}, switching off the supervision of $m$ and $\alpha$ leads to a significant performance drop of 0.89dB, compared with SRUDC-full. We further try to justify the effect of the whole scattering branch, by deleting it from SRUDC. In this case, SRUDC degenerates to a variant of Unet. As can be seen from the second row of Table \ref{tab:scattering_branch}, the whole scattering branch brings a significant performance gain of around 1.36dB PSNR. In addition, we substitute TSABs in the scattering branch by CNN structure GCABs to show the effectiveness of the CSA. It can be observed that our proposed TSABs could introduce a significant performance improvement of around 1dB PSNR gain, due to the strong capability of modeling global information.


\noindent\textbf{\underline{Image branch:}} We evaluate the components in the proposed GCAB in the image branch. We sequentially remove the gating operation and channel attention and retrain the counterpart models. As shown in Table \ref{tab:image_branch}, eliminating the gating operation and the channel attention respectively causes 0.38 dB and 0.7 dB PSNR drops.
\begin{table}[t]
	\centering
	\caption{Effect of the components in the image branch.}
	\label{tab:image_branch}
	\begin{small}
		\begin{tabular}{l|ccc}
			\toprule[1pt]
			Model & PSNR & SSIM  & LPIPS \\
			\midrule
			w/o gating operation & 35.64 & 0.9824 & 0.0356 \\
			w/o channel attention & 35.32 &  0.9806 & 0.0417 \\
			GCAB & \textbf{36.02} & \textbf{0.9833} & \textbf{0.0322}\\
			\bottomrule[1pt]
		\end{tabular}
	\end{small}
 \vspace{-3mm}
\end{table}

\begin{table}[t]
	\centering
	\caption{Effect of the feature fusion block.}
	\label{tab:feature_fusion block}
	\begin{small}
		\begin{tabular}{l|ccc}
			\toprule[1pt]
			Model & PSNR & SSIM  & LPIPS \\
			\midrule
			Concatenation & 35.84 & 0.9813 & 0.0325 \\
			SKfusion \cite{li2019selective} & 35.42  & 0.9824 & 0.0335 \\
			SCAM \cite{Song2023R4} & 33.59  & 0.9707 & 0.0479 \\
			Our FFB & \textbf{36.02} & \textbf{0.9833} & \textbf{0.0322}\\
			\bottomrule[1pt]
		\end{tabular}
	\end{small}
 \vspace{-5mm}
\end{table}

\begin{figure}[t!]
	\centering
	\subfigure[Real UDC input]{
		\includegraphics[width=0.325\linewidth]{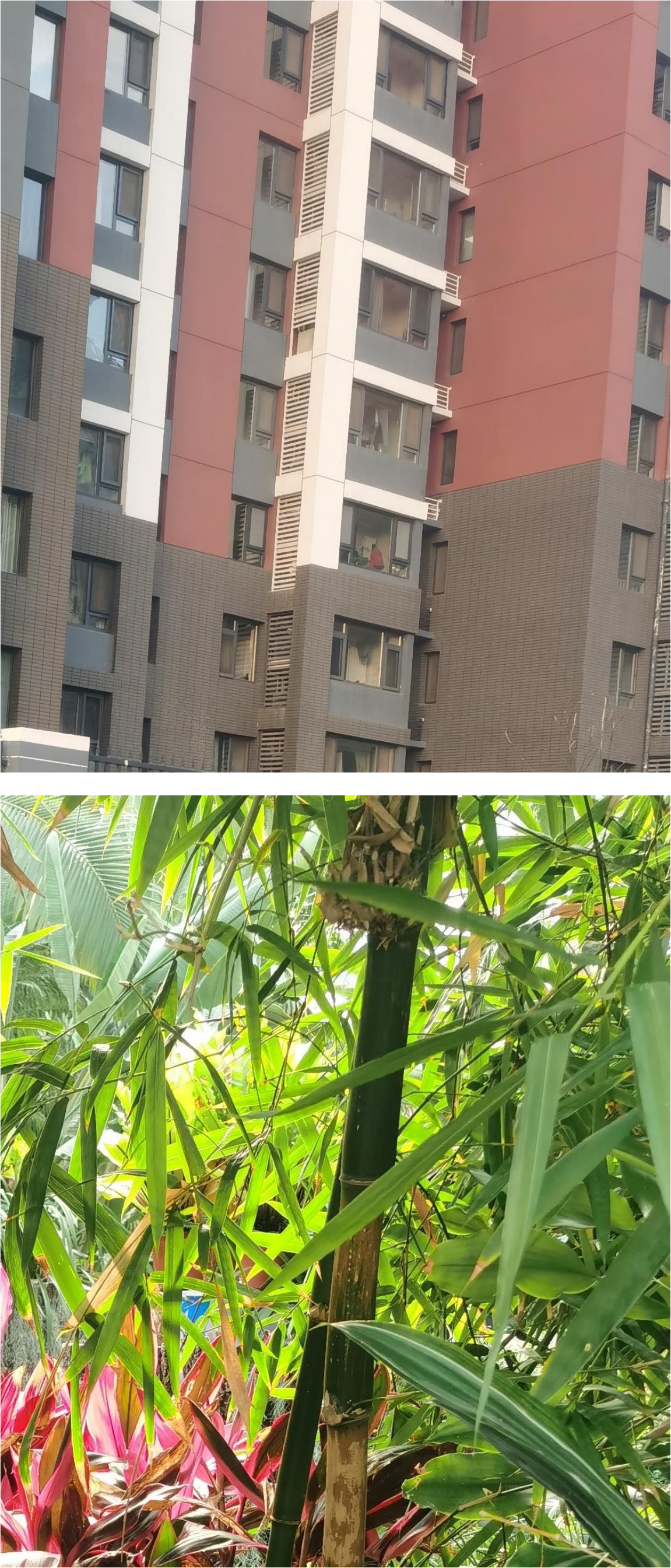}}
	\hspace{-2mm}
	\subfigure[Pretrained]{
		\includegraphics[width=0.325\linewidth]{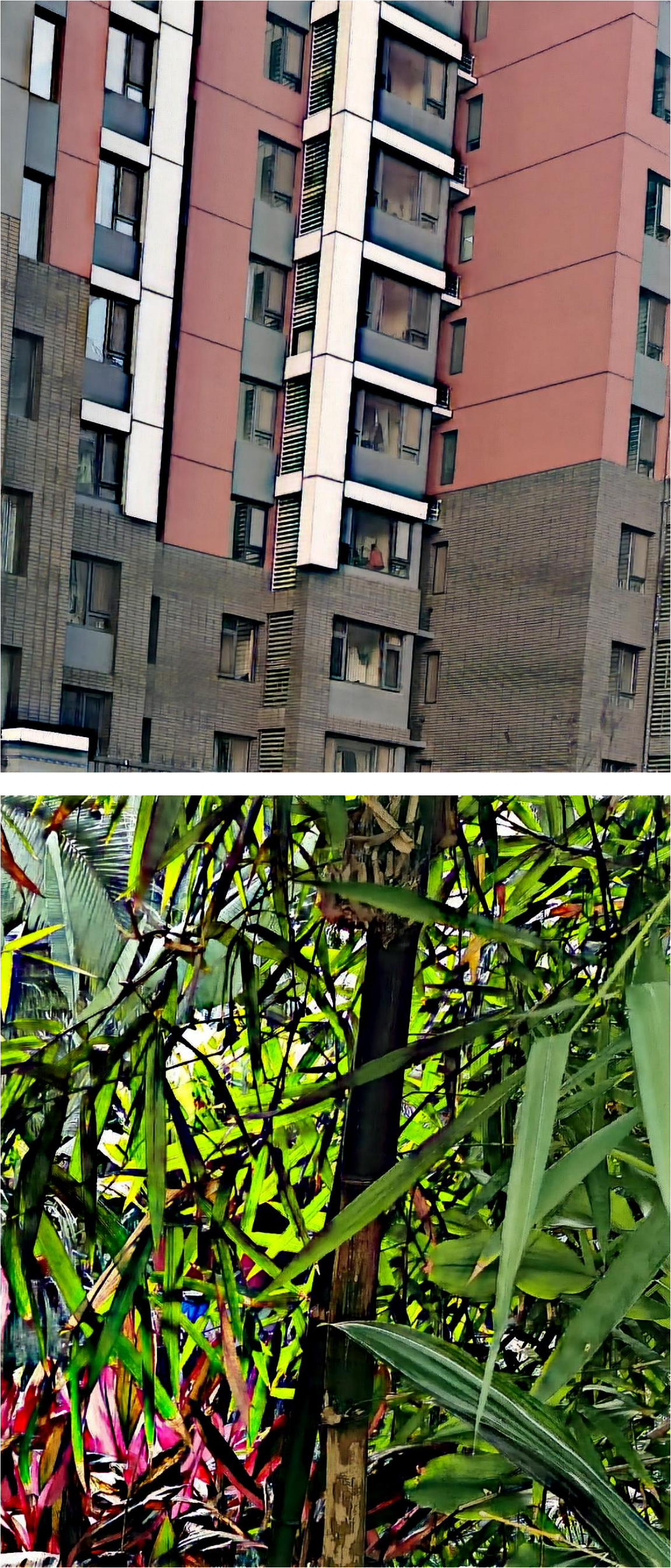}}
	\hspace{-2mm}
	\subfigure[Retrained]{
		\includegraphics[width=0.325\linewidth]{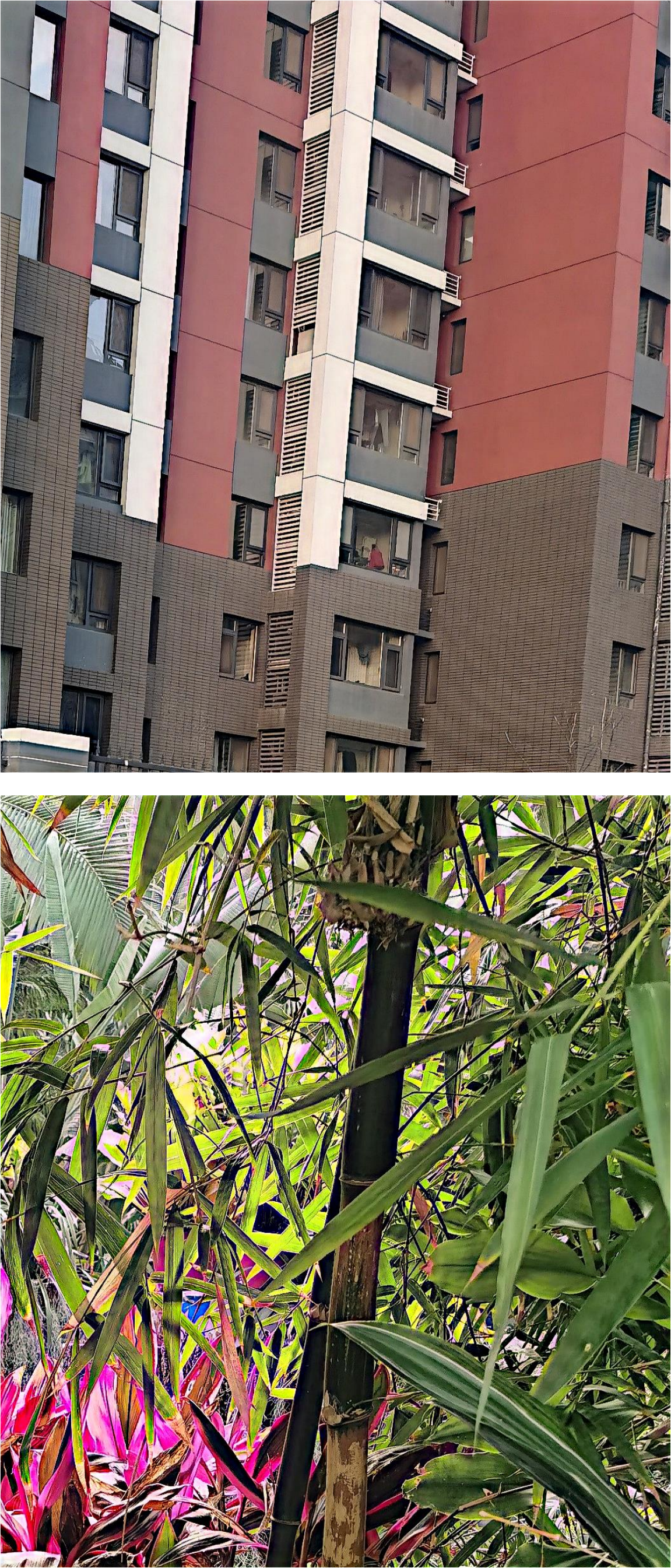}}
	\hspace{-2mm}
	\caption{Comparison of BNUDC pretrained and retrained with synthetic data from different IFPs. (a) Real-world input UDC images. (b) Results of BNUDC pretrained with data generated by Feng \textit{et al.} \cite{feng2021removing}. (c) Results of BNUDC retrained with data from our enhanced IFP. Our synthetic data effectively improves the visual quality of the restored results from BNUDC.}
	\label{fig:result_IFP}
 \vspace{-4mm}
\end{figure}

\noindent\textbf{\underline{Feature Fusion Block:}} To show the benefit of the FFB, we build an ablation experiment on this module. Specifically, we substitute FFB with the concatenation, SKfusion \cite{li2019selective} and SCAM \cite{Song2023R4}. The results summarized in Table \ref{tab:feature_fusion block} verify the superiority of SFT with around 0.18dB PSNR gains.

\subsection{Effectiveness of Enhanced IFP}
To present the importance of considering the scattering effect in the UDC image restoration, we compare the pretrained and retrained model of the state-of-the-art UDC restoration method BNUDC. The pretrained model is trained with the data generated from the existing IFP, Feng \textit{et al.} \cite{feng2021removing}, and the retrained model is trained with data synthesized by our enhanced IFP. The visual results tested on the real-world UDC images are shown in Fig. \ref{fig:result_IFP}. From the comparison, our synthetic data effectively improves the visual quality of the restored results from BNUDC.


\section{Conclusion} \label{sec:conclusion}
In this study, we propose to address the UDC image restoration problem with the consideration of the scattering effect caused by the display. With the physical UDC scattering model, we improve the IFP for the realistic UDC image synthesis. To recover the clean background scenes, we specially design a dual-branch deep network, where the scattering branch estimates the parameters of the scattering effect from degraded images, while the image branch restores the background image. The devised feature fusion block provides guidance to the UDC image restoration with the global information from the scattering branch. Our model significantly outperforms the state-of-the-art methods on both synthetic and real-world UDC images.

\indent \textbf{Acknowledgement.} This work was supported  in part by Macau Science and Technology Development Fund under SKLIOTSC-2021-2023, 0072/2020/AMJ, and 0022/2022/A1; in part by Research Committee at University of Macau under MYRG2020-00101-FST and MYRG2022-00152-FST; in part by Natural Science Foundation of China under 61971476; and in part by Alibaba Group through Alibaba Innovative Research Program.

{\small
\bibliographystyle{ieee_fullname}
\bibliography{SRUDC}
}

\end{document}